\documentclass[lettersize,journal]{IEEEtran}
\usepackage{amsmath,amsfonts}
\usepackage{algorithm}
\usepackage{array}
\usepackage[caption=false,font=normalsize,labelfont=sf,textfont=sf]{subfig}
\usepackage{textcomp}
\usepackage{stfloats}
\usepackage{url}
\usepackage{verbatim}
\usepackage{graphicx}
\usepackage{cite}
\usepackage{hhline}
\usepackage{amssymb}
\usepackage{multirow}
\usepackage{color}
\usepackage{colortbl}
\usepackage{hhline}
\usepackage{booktabs}
\usepackage{amssymb}
\usepackage{calligra}
\usepackage{algorithm}
\usepackage{algorithmicx}
\usepackage{algpseudocode} 
\usepackage{amsmath} 
\usepackage{algorithm}
\errorcontextlines\maxdimen
\makeatletter
\newcommand*{\algrule}[1][\algorithmicindent]{\makebox[#1][l]{\hspace*{.5em}\thealgruleextra\vrule height \thealgruleheight depth \thealgruledepth}}%
\newcommand*{\thealgruleextra}{}
\newcommand*{\thealgruleheight}{.75\baselineskip}
\newcommand*{\thealgruledepth}{.25\baselineskip}
\newcount\ALG@printindent@tempcnta
\def\ALG@printindent{%
	\ifnum \theALG@nested>0
	\ifx\ALG@text\ALG@x@notext
	\else
	\unskip
	\addvspace{-1pt}
	\ALG@printindent@tempcnta=1
	\loop
	\algrule[\csname ALG@ind@\the\ALG@printindent@tempcnta\endcsname]%
	\advance \ALG@printindent@tempcnta 1
	\ifnum \ALG@printindent@tempcnta<\numexpr\theALG@nested+1\relax
	\repeat
	\fi
	\fi
}%
\usepackage{etoolbox}
\patchcmd{\ALG@doentity}{\noindent\hskip\ALG@tlm}{\ALG@printindent}{}{\errmessage{failed to patch}}
\makeatother

\newbox\statebox
\newcommand{\myState}[1]{%
	\setbox\statebox=\vbox{#1}%
	\edef\thealgruleheight{\dimexpr \the\ht\statebox+1pt\relax}%
	\edef\thealgruledepth{\dimexpr \the\dp\statebox+1pt\relax}%
	\ifdim\thealgruleheight<.75\baselineskip
	\def\thealgruleheight{\dimexpr .75\baselineskip+1pt\relax}%
	\fi
	\ifdim\thealgruledepth<.25\baselineskip
	\def\thealgruledepth{\dimexpr .25\baselineskip+1pt\relax}%
	\fi
	\State #1%
	\def\thealgruleheight{\dimexpr .75\baselineskip+1pt\relax}%
	\def\thealgruledepth{\dimexpr .25\baselineskip+1pt\relax}%
}

\begin{document}

\title{Dynamic Enhancement Network for Partial Multi-modality Person Re-identification}

\author{
		Aihua Zheng,
	    Ziling He,
		Zi Wang,
		Chenglong Li*,
		Jin Tang
		
		\thanks{
			This research is supported in part by the National Natural Science Foundation of China (61976002), the Natural Science Foundation of Anhui Higher Education Institution of China (KJ2020A0033), and the University Synergy Innovation Program of Anhui Province (GXXT-2021-038 and GXXT-2019-025).}
			
		\thanks{
			A. Zheng and C. Li are with the Information Materials and Intelligent Sensing Laboratory of Anhui Province, Anhui Provincial Key Laboratory of Multimodal Cognitive Computation, School of Artificial Intelligence, Anhui University, Hefei, 230601, China
			(e-mail: ahzheng214@foxmail.com;  lcl1314@foxmail.com).}
			
			\thanks{Z. He, Z. Wang and J. Tang are with Anhui Provincial Key Laboratory of Multimodal Cognitive Computation, School of Computer Science and Technology, Anhui University, Hefei, 230601, China (e-mail: zlhe0605@foxmail.com; ziwang1121@foxmail.com; tangjin@ahu.edu.cn)
		}
	}


\markboth{IEEE TRANSACTIONS ON NEURAL NETWORKS AND LEARNING SYSTEMS}%
{Zheng \MakeLowercase{\textit{et al.}}: Dynamic Enhancement Network for Partial Multi-modality Person Re-identification}


\maketitle

\begin{abstract}
Many existing multi-modality studies are based on the assumption of modality integrity. However, the problem of missing arbitrary modalities is very common in real life, and this problem is less studied, but actually important in the task of multi-modality person re-identification (Re-ID). To this end, we design a novel dynamic enhancement network (DENet), which allows missing arbitrary modalities while maintaining the representation ability of multiple modalities, for partial multi-modality person Re-ID. To be specific, the multi-modal representation of the RGB, near-infrared (NIR) and thermal-infrared (TIR) images is learned by three branches, in which the information of missing modalities is recovered by the feature transformation module. Since the missing state might be changeable, we design a dynamic enhancement module, which dynamically enhances modality features according to the missing state in an adaptive manner, to improve the multi-modality representation. Extensive experiments on multi-modality person Re-ID dataset RGBNT201 and vehicle Re-ID dataset RGBNT100 comparing to the state-of-the-art methods verify the effectiveness of our method in complex and changeable environments.
\end{abstract}



\section{Introduction}
\IEEEPARstart{A}{fter} continuous development, person re-identification (Re-ID) has attracted high attention with great achievement recently. However, conventional Re-ID methods only rely on single visible images, which results in limited application in low illumination scenarios such as hazy or dark environment. 
To facilitate the Re-ID task in low illumination scenario, Wu \textit{et al.} \cite{Wu2017RGBIR} first launch the RGB and near-infrared cross-modality Re-ID. Despite of the recent efforts in cross-modality Re-ID, the additional heterogeneity between the visible and infrared modalities brings huge challenging for cross-modality Re-ID.

With the popularity of diverse kinds of cameras (e.g., infrared cameras), multi-modality person re-identification and vehicle re-identification receives increasing interests in the computer vision community~\cite{2013Tri,Zheng2021Multi}, due to the strong complementary benefits from different modalities.
Li \textit{et al.}~\cite{Li2020HAMNet} first contribute the multi-spectral vehicle datasets RGBN300 and RGBNT100, and propose a heterogeneity-collaboration aware multi-stream revolutionary network for multi-spectral vehicle Re-ID task.
Recently, Zheng \textit{et al.} \cite{Zheng2021Multi} contribute a multi-modality Re-ID benchmark dataset RGBNT201 containing images of each person in three modality: visible, near-infrared and thermal-infrared, making full use of the complementary information of multiple modalities. Meanwhile, they propose a progressive fusion method to combine multi-modality information.

\begin{figure}
\begin{center}
\includegraphics[width=0.98\linewidth]{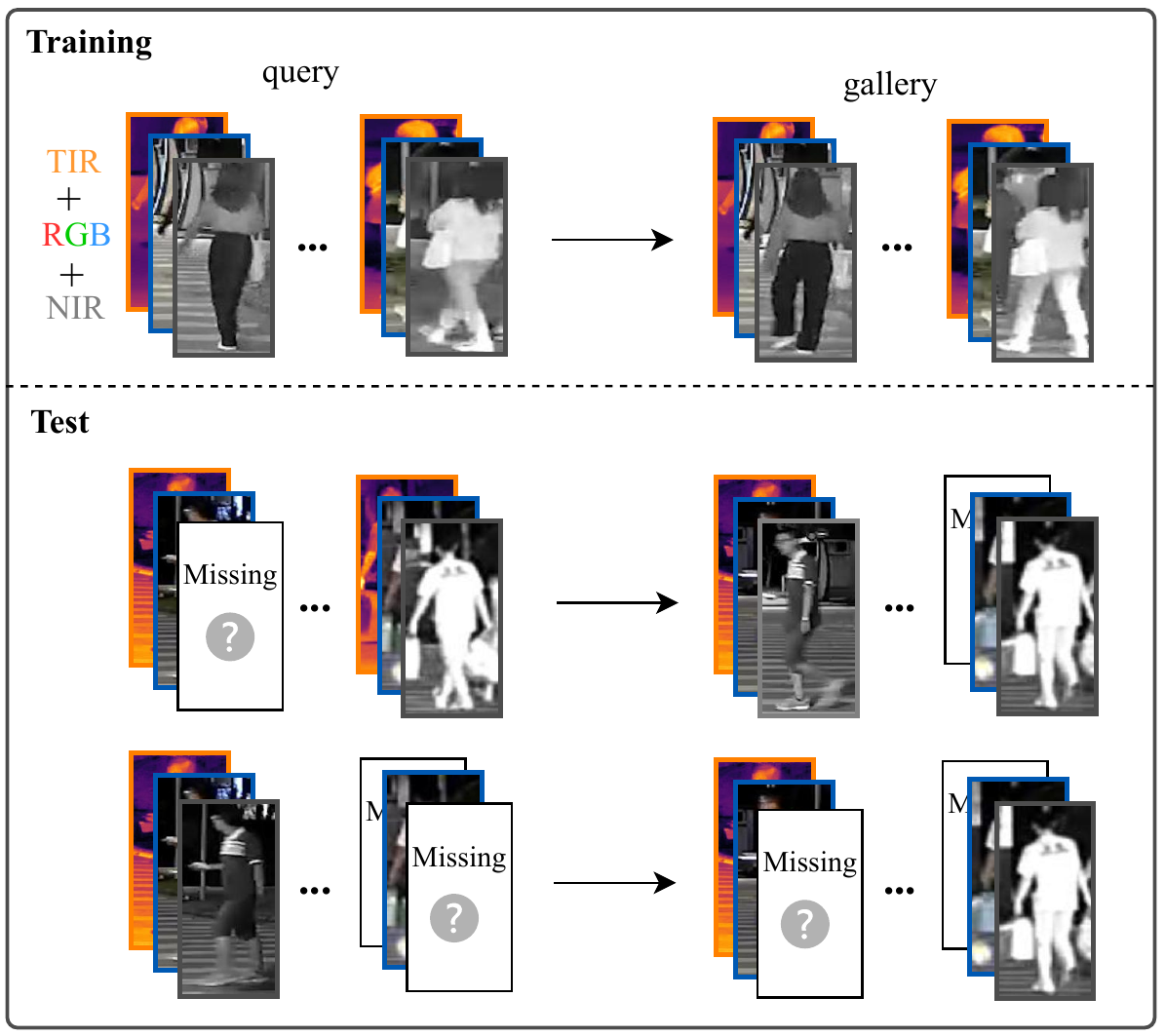}
\end{center}
\caption{The training data is complete and multi-modality, but the image may lose one or two modalities during the test stage.}
\label{fig:motivation}
\end{figure}

Although multi-modality images have great advantages, the requirements for complete multi-modality data are relatively strict. As shown in Fig.~\ref{fig:motivation}, one or two modalities may arbitrarily-missing during the test stage caused by shooting conditions, equipment damage or storage errors. Therefore, partial multi-modality Re-ID is essential in real-life applications.


\setlength{\abovecaptionskip}{2pt} 
\setlength{\belowcaptionskip}{-2pt}


There are three major issues to be addressed in partial multi-modality Re-ID. 
\begin{itemize}
    \item The missing state of modalities is arbitrary. Traditional methods usually train an independent model for each situation, which is not only inefficient, but also low scalability.
    
    \item The recovery of missing modalities is a difficult task. Existing generative adversarial network (GAN) based methods \cite{2018Translating,2017Unsupervised} try to use existing data to generate missing data. However, the large amount of training data and unstable training process significantly limit the performance.
    
    \item The ability of multi-modality representation is affected when some modalities are missing. Existing methods often enhance feature representation through multi-modality fusion~\cite{Marco2014Multimodal,2014Medical,2017Deep}. But direct and fixed fusion approaches lose modality-specific information and can not adapt to a variety of missing states.
\end{itemize}

To handle the first issue, we design a novel dynamic enhancement network (DENet) to automatically recover the representation of missing modalities and perform dynamic interactions according the missing state. In particular, DENet contains three feature extraction branches and a set of feature transformation branches. The former is used to extract features of available modalities, and the latter is to recover features of missing modalities. These branches only need training once, and then dynamically compose the feature representation and perform dynamic enhancement according to the missing state in the test stage.

To solve the second issue, we propose the cross-modality feature transformation (CMFT), including the up-sample and down-sample structures. Specifically, we send the features of available modality into CMFT to transform it to the features of the missing modality. The missing image recovered by up-sample operation is constrained by reconstruction loss, and then the output representations of CMFTs are constrained by the similarity loss, so as to recover more real and discriminative missing information. In the training stage, we train the CMFT for each feature transformation between one modality to another one using the multi-modality data. In the testing stage, we dynamically use the CMFT moudle to handle the arbitrary missing state.

Finally, for the third issue, we propose a dynamic enhancement module (DEM). The DEM gets rid of the bondage of fixed fusion and adopts dynamic cutting strategy to realize the feature enhancement of arbitrary missing cases, so as to process and correlate the information of multiple modalities in different missing states. Specifically, we first build a complete directed enhancement graph by taking the modalities as graph nodes, and then dynamically cut some graph edges according the missing state. In particular, if a modality is missing, we cut two edges that take this node as the arc tail. Based on the generated graph, we achieve the feature enhancement under the missing state. Through this way, we can adapt to a variety of missing states in a dynamic manner.

Our contributions are summarized as follows.
\begin{itemize}
    \item We propose a novel dynamic enhancement network to solve the problem of missing modalities which frequently occur in real-world scenario in multi-modality Re-ID task.
    
    \item To recover the missing data, we design the cross-modality feature transformation module with the constraints of the reconstruction loss and similarity loss.
    
    \item To achieve feature enhancement for arbitrary missing states, we propose a dynamic cutting strategy to adjust the enhancement graph adaptively.
    
    \item We carry out extensive comparative experiments with different state-of-the-art methods on RGBNT201 and RGBNT100 to verify the effectiveness of DENet in various missing cases.
    
\end{itemize}


\section{Related Work}

\subsection{Person Re-ID}
Person re-identification mainly solves the recognition and matching problem of person images taken by different non overlapping cameras \cite{Ye2018dual,Luo2019strong,2019sdml-TNNLS,2019JLDMFS-TNNLS,2020Relation,2022SR-TNNLS}. It can identify person according to their wearing, posture, hairstyle and other information, so it is widely used in intelligent security and other fields. 
As one of the hot topics in computer vision research, it has made great progress after years of development. And these deep learning-based Re-ID methods can be roughly divided into two categories: feature learning and distance metric learning. The feature learning network aims to learn a robust and discriminative feature representation for person image. For example, Wei \textit{et al.} \cite{GLAD} proposed a global-local-alignment descriptor (GLAD) to overcome pose changes and misalignments. Metric learning aims to learn the similarity between images. The more commonly used metric learning loss is the triple loss \cite{Triplet-loss}, and there are continuous works to improve this loss function to significantly improve the Re-ID performance \cite{Improved_Triplet_Loss,Beyond_triplet_loss}.
At the same time, it also faces great challenges, such as different viewpoints \cite{2020Viewpoint}, illumination changes \cite{ACM-MM_Illumination-Invariant}, occlusion \cite{CVPR_occlusion-free}, etc. Especially in low light conditions, the RGB single-modality Re-ID can not adapt to the night environment, so its performance is limited.


\subsection{Cross-Modality and Multi-Modality Re-ID}
To overcome the illumination challenges of the RGB single-modality Re-ID task, Wu \textit{et al.} \cite{Wu2017RGBIR} first launch the RGB-infrared cross-modality person re-identification task with the infrared and visible cross-modality person Re-ID dataset SYSU-MM01 and a baseline method Deep Zero-Padding. 
The emerging achievements of this task roughly fail into the following three categories: 1) Representation learning based methods, which aim to extract the robust and discriminative features shared by two modality images \cite{Ye2018,Lu2020SSFT}. 2) Metric learning based methods, the key of which kind of methods is to design a reasonable metric method or loss function to learn the similarity of two images \cite{Ye2018dual,HCloss}. 3) Modality transformation based methods, which transform the cross-modality task into the single-modality task, so as to reduce the modality heterogeneity \cite{Dai2018Cross,dual2019}. 

In order to overcome the limitation in RGB Re-ID task as well as the heterogeneity problem in cross-modality Re-ID task, the multi-modality Re-ID task is proposed. Mogelmose \textit{et al.} propose a tri-modal person Re-ID\cite{2013Tri}, combining RGB, depth and thermal data. Special features are obtained from three modalities, which are then evaluated using a joint classifier. Zheng \textit{et al.} \cite{Zheng2021Multi} take advantage of the complementary strengths of the visible, near-infrared and thermal-infrared modalities to achieve robust Re-ID performance. This work contributes a comprehensive benchmark dataset RGBNT201, including 201 identities captured from various challenging conditions, to facilitate research on RGB-NIR-TIR multi-modality person re-identification.


\begin{figure*}
\setlength{\abovecaptionskip}{2pt} 
\setlength{\belowcaptionskip}{-2pt} 
\begin{center}
\includegraphics[width=1\linewidth]{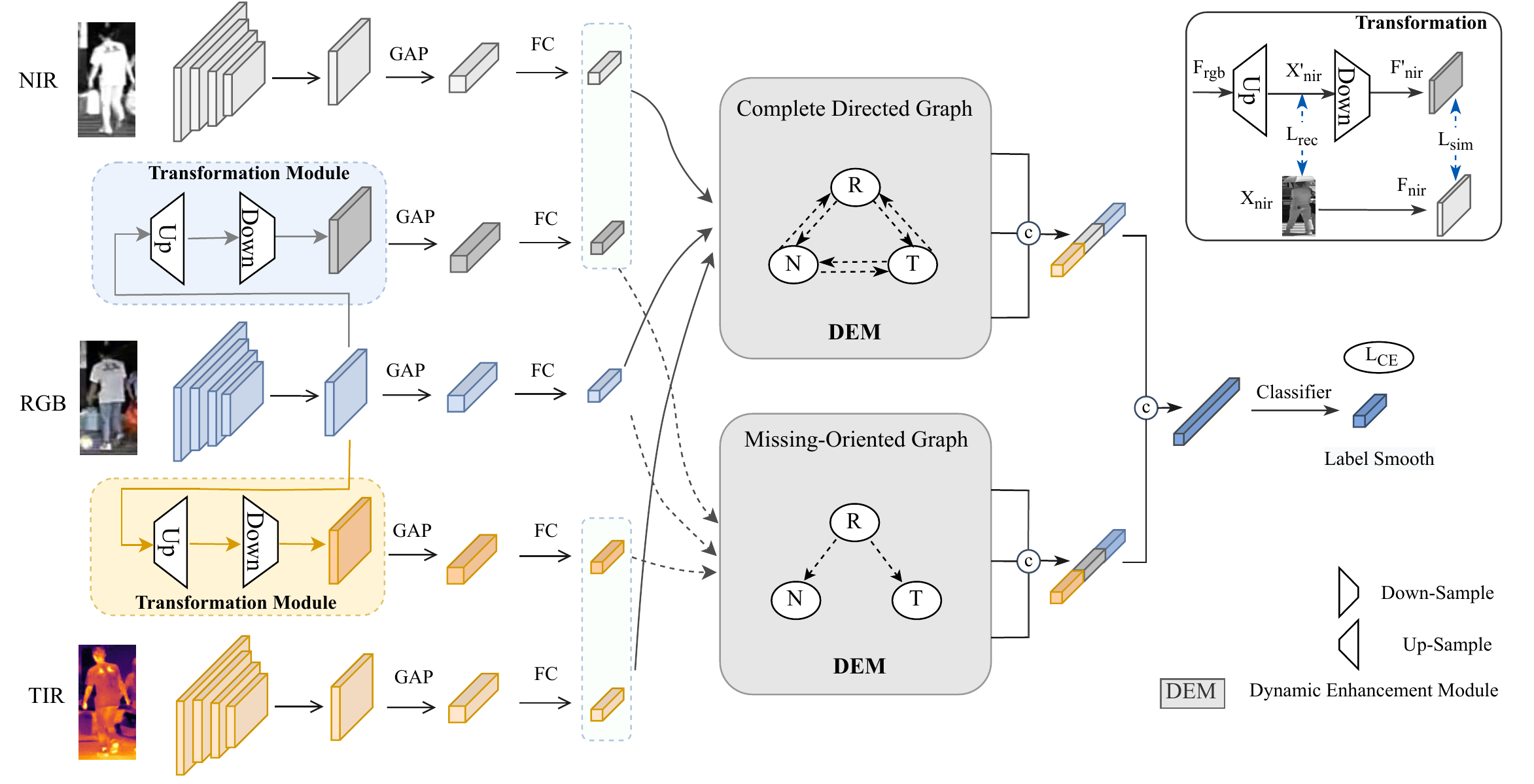}
\end{center}
\caption{
An overview of our proposed partial multi-modality re-identification network DENet. First, we extract the features of each modality from individual target image triplet. To solve the missing modalities, we employ the the cross-modality feature transformation (CMFT), including the up-sample and down-sample structures. And then, we feed the obtained features into the dynamic enhancement module(DEM), which can be adjusted according to different arbitrarily-missing scenarios to maintain the representation ability of multiple modalities. Finally, we concatenate the enhanced features to obtain the final representation.
}
\label{fig:overview}
\end{figure*}

\subsection{Partial Multi-Modality Learning}
In view of the wide application of multi-modality learning, many fields are committed to the exploration of this study, and some work have paid attention to the problem of missing modality. When processing medical images, Zhang \textit{et al.} \cite{Zhang2018Translating} propose to use CycleGAN to generate missing information at the data level. Secondly, based on the imputation method, HeMIS \cite{Havaei2016HeMIS} use statistical features as embedding for decoding, and feature fusion adopts the fusion calculation method of mean and variance. In addition, Shen \textit{et al.} \cite{Shen2019} propose an adaptive network to design a loss function for the model to generate features similar to the real features in the case of modality-missing. Tsai \textit{et al.} \cite{Tsai2019representation} point out in their work on multi-modality representation learning that models must be robust to unexpected missing or noisy modalities during testing, and they propose to optimize for a joint generative-discriminative objective across multi-modality data and labels.

\subsection{Multi-Modality Feature Representation and Fusion}
 Multi-modality feature representation plays an important role in multi-modality tasks. The main task is to learn better feature representation of multi-modality data with the complementarity of multiple modalities. The main problem is how to combine the data from different modalities with different degrees of noise. 
 Two commonly used multi-modality representations are the joint representation and the coordinated representation.
 The former maps the information of different modalities to the same feature space \cite{2018TPAMI}, and the later maps the information of each modal respectively with certain constraints between each modality after mapping \cite{2018Trans}.
Another key issue is the multi-modality feature fusion, which integrates the information from different modalities through different fusion strategies to obtain more discriminative information for specific tasks. 
%
%
It can be roughly divided into early fusion \cite{Poria-multimodal}, late fusion \cite{FSSD,DenseNet}, and mixed fusion \cite{DoubleFusion}.


\section{DENet: Dynamic Enhancement Network}
In order to solve the ubiquitous modality-missing problem in multi-modality Re-ID task, we propose the Dynamic Enhancement Network (DENet), which consists of multi-branch feature extraction, cross-modality feature transformation module, and dynamic enhancement module, as shown in Fig.~\ref{fig:overview}. In the following, we describe their details.

\subsection{Multi-branch Feature Extraction}
In training phase, we use all real visible, near-infrared, and thermal-infrared modalities. In order to obtain complementary information, we design a multi-branch network to extract the features of each modality from individual image triplet. We select the commonly used ResNet50 \cite{He2016Residual} as the backbone. Furthermore, we add the channel and spatial attention layer \cite{2018CBAM} to each branch to focus the most informative features. It is worth noting that the three ResNet50 branches are independent to each other without sharing the parameters. 
The feature $F_{rgb}$, $F_{nir}$ and $F_{tir}$ extracted by corresponding branch preserve modality-specific information as much as possible.

\subsection{Cross-modality Feature Transformation Module}
In order to deal with unpredictable modality-missing problem in the testing phase, we introduce the cross-modal feature transformation (CMFT) to learn the conversion relationship between the existing modality and missing modality, as shown in Fig.~\ref{fig:overview}. Taking the example that NIR and TIR modalities are missing, we use two CMFTs, specifically called R2N and R2T and take the R2N as an example to illustrate the details. 

The module passes the RGB feature $F_{rgb}$ through a up-sample block to obtain the intermediate fake-image and then obtains the recover feature through down-sample block.
\begin{equation}
\begin{aligned}
X_{nir}^{'} = UpSample(F_{rgb}),
\end{aligned}
\end{equation}

\begin{equation}
\begin{aligned}
F_{nir}^{'} = DownSample(X_{nir}^{'}).
\end{aligned}
\end{equation}

We add pixel-level constraints to above images and features by minimizing
\begin{equation}
\begin{aligned}
L_{rec} = \|X_{nir}-X_{nir}^{'}\|_{2}^{2},
\end{aligned}
\end{equation}

\begin{equation}
\begin{aligned}
L_{sim}=\|\theta_{n}(G(F_{nir}))-\theta_{n}^{'}(G(F_{nir}^{'}))\|_{1}.
\end{aligned}
\end{equation}
where $X_{nir}$ denotes NIR images, $G$ denotes Global Average Pooling, $\theta_{n}(x)=ReLU(BN(W_{n}x))$ and $\theta_{n}^{'}(x)=ReLU(BN(W_{n}^{'}x))$ denote the projection functions that embed the feature vectors into the same space. The two functions are equivalent to provide an effective supervision for the transformation module to obtain more adequate semantic information.

The CMFT is trained end-to-end, and fixed all parameters after training, then works in testing phase to recover more realistic and discriminative missing information under missing scenario.


\begin{figure}[t]
\setlength{\abovecaptionskip}{2pt} 
\setlength{\belowcaptionskip}{-2pt} 
\begin{center}
\includegraphics[width=0.98\linewidth]{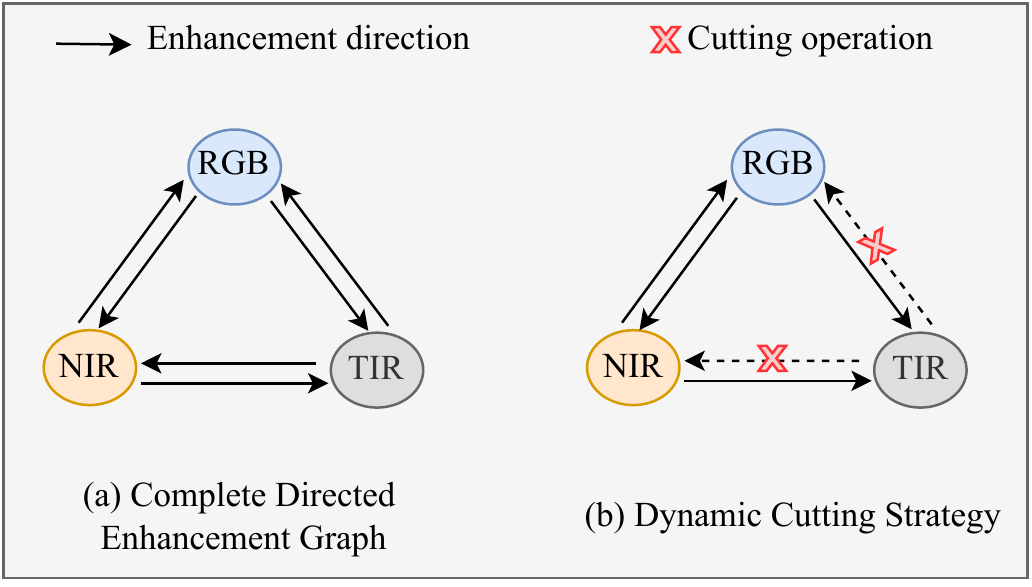}
\end{center}
\caption{
(a) The complete enhancement graph. (b) The dynamic-cut strategy taking the missing of TIR as an example.
}
\label{fig:dynamic}
\end{figure}


\begin{figure}[t]
\setlength{\abovecaptionskip}{2pt} 
\setlength{\belowcaptionskip}{-2pt} 
\begin{center}
\includegraphics[width=0.98\linewidth]{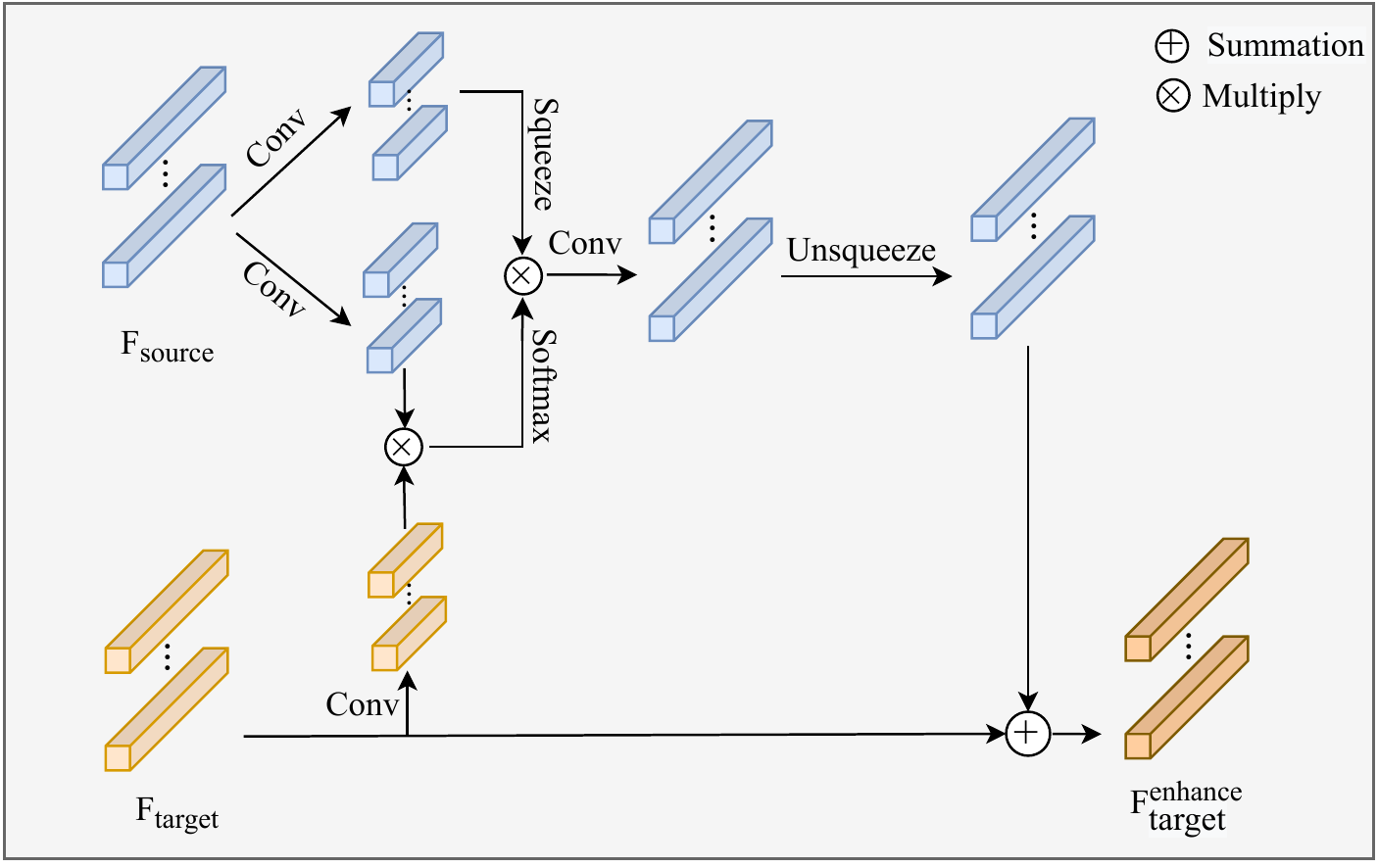}
\end{center}
\caption{
The detail of the enhancement operation. Where the feature of target modal is enhanced by the source feature to obtain the corresponding improved feature.
}
\label{fig:enhancement}
\end{figure}


\subsection{Dynamic Enhancement Module}
Each individual branch is concerned with the learning of discriminative features in a given deterministic modality. For the purpose to make a single branch use the internal correlation between different modalities  without interactive information, we adopt the enhancement operation. However, considering the unpredictable missing modalities, how to maintain the ability of multi-modality representation is crucial. Existing enhancement methods with fixed structure \cite{CVPR_AMC,2020MMCA} are not suitable for our problem. To this end, we propose the Dynamic Enhancement Module (DEM), which can adaptively adjust the enhancement branches according to the missing states to maintain the representation ability of multiple modalities.

As shown in Fig.~\ref{fig:dynamic} (a), we construct a complete directed enhancement graph by taking the modalities as graph nodes, and we show the dynamic cutting strategy in the case of missing TIR in Fig.~\ref{fig:dynamic} (b), which cuts the two enhancement edges that take the TIR node as the arc tail adaptively. The details of cross-modality enhancement are shown in Fig.~\ref{fig:enhancement}, in which the feature $F_{target}$ of the target modal is enhanced by the source feature $F_{source}$ to obtain the corresponding $F_{target}^{enhance}$. This progress can be described below. First, we implement three 1×1 convolutions of $F_{source}$ and $F_{target}$ to obtain $F_{source}^{'}$, $F_{source}^{''}$ and $F_{target}^{'}$. Then we perform dot product as follows,

\begin{equation}
\begin{aligned}
F_{source}^{'''} = ( F_{source}^{'} \otimes F_{target}^{'} ) \otimes F_{source}^{''}.
\end{aligned}
\end{equation}
After convolution and unsequeeze operations, we obtain the final representation $F_{target}^{enhance}$ by adding the original target feature,
\begin{equation}
\begin{aligned}
F_{target}^{enhance} = F_{source}^{'''} + F_{target}.
\end{aligned}
\end{equation}
At last, we integrate different modalities by concatenating corresponding features to form the final person or vehicle discriminator.
\begin{equation}
\begin{aligned}
F_{c1} = Concat(F_{rgb}^{enhance}, F_{nir}^{enhance}, F_{tir}^{enhance}),
\end{aligned}
\end{equation}

\begin{equation}
\begin{aligned}
F_{c2} = Concat(F_{rgb},F_{nir'}^{enhance},F_{tir'}^{enhance}),
\end{aligned}
\end{equation}

\begin{equation}
\begin{aligned}
F_{final} = Concat(F_{c1},F_{c2}).
\end{aligned}
\end{equation}

In the test stage, if there is no missing (i.e. missing rate $\eta$ = 0), then the model obtains the discriminant features as usual. If missing, we can judge the missing modalities and missing rate $\eta$, and complete the missing information through our CMFT module, so as to get different feature combinations under different missing states to adapt to the missing situation.


\subsection{Objective Function}
Since the features obtained after global average pool (GAP) are in Euclidean space and the triplet loss is suitable for constraints in free Euclidean space, we apply the triplet loss directly to the features after GAP. Then, the features are normalized to the hypersphere through BN layer to optimize the classification loss. The triplet loss and cros-entropy loss (CE loss) is computed as:
\begin{equation}
\begin{aligned}
L_{tri}=max \{ d(a,p)-d(a,n)+\alpha, 0 \},
\end{aligned}
\end{equation}

\begin{equation}
\begin{aligned}
L_{CE} = - \sum_{i=1}^{N} q_{i}log(p_{i})
\begin{cases}
q_{i}=0 ,y \neq i  \\
q_{i}=1 ,y = i,
\end{cases}
\end{aligned}
\end{equation}
where $d(a,p)$ and $d(a,n)$ are feature distances of positive and negative pairs and the $\alpha$ is the margin of triplet loss.

In order to prevent the network from overfitting, we adopt the label smooth strategy \cite{Inceptionv2}. Label smooth will change the real probability distribution and cross-entropy loss function as follows,
\begin{equation}
\begin{aligned}
p_{i}=
\begin{cases}
1-\beta &  ,i = y   \\
\beta / N &   ,i \neq y,
\end{cases}
\end{aligned}
\end{equation}

\begin{equation}
\begin{aligned}
L_{CE}^{'}=
\begin{cases}
(1-\beta) \ast L_{CE} &   ,i = y \\
\beta \ast L_{CE} &   ,i \neq y,
\end{cases}
\end{aligned}
\end{equation}
where $\beta$ is a small hyperparameters, which encourage the model to be less confident on the training set.

Through the above description, the overall loss function in the training phase can be formulated as:
\begin{equation}
\begin{aligned}
L=L_{ReID} + \rho L_{rec}+ \mu L_{sim},
\end{aligned}
\end{equation}
where $L_{ReID}$ denotes all the triplet loss and cross entropy loss we used. $\rho$ and $\mu$ are the balance hyperparameters between these losses. We set $\rho=1$ and $\mu=1$ in training stage according to the experiment results of parameter analysis.

\subsection{Implementation Details}
The implementation platform is Pytorch with a NVIDIA GTX 1080Ti GPU. We use the ResNet-50 pre-trained on ImageNet as the backbone to extract 2,048d features for RGB, NIR and TIR images from the global average pooling layer. The batch size is set to 8 and the initial learning rate is set to 0.01 and then reduce it to 0.001 and 0.0001 at epoch 30 and 55. In the training phase, we set the dropout to 0.5 to prevent overfitting. And we use stochastic gradient descent (SGD), setting momentum to 0.9 and weight decay to 0.0005 to fine-tune the network.


\section{Evaluation}  
To verify the effectiveness of our proposed method, we compare it with state-of-the-art methods on the benchmark multi-modality person Re-ID dataset RGBNT201 (Zheng \textit{et al.}, 2021) and multi-spectral vehicle dataset RGBNT100 (Li \textit{et al.}, 2020).

\subsection{Datasets and Evaluation Protocols}
\noindent
\textbf{RGBNT201} \cite{Zheng2021Multi} is the first multi-modality person Re-ID dataset with four non-overlapping views collected in a campus scene, each with three cameras recording RGB, NIR, and TIR data simultaneously. It covers the challenges of part occlusion, low illumination, background clutter, high illumination, motion blur, low illumination and so on. The dataset contains 201 identities, and we select 141 identities for training, 30 identities for verification, and the remaining 30 identities for testing.

\noindent
\textbf{RGBNT100} \cite{Li2020HAMNet} is a multi-spectral vehicle Re-ID dataset contributed by Li \textit{et al.} in 2020. It contains RGB, NIR and TIR vehicle images of 100 identities from 8 camera views. The training set in RGBNT100 contains 8675 image triples of 50 vehicles. The other 50 vehicles with 8575 image triples are used for the test, and 1715 image triples are randomly selected as the query set.

\noindent
\textbf{Evaluation Protocols}. Followed by the evaluation protocols in the commonly used Market-1501 dataset, we use mean Average Precision (mAP) and Cumulative Matching Characteristics (CMC) as the metrics. mAP is the area under the precision recall curve, which measures the quality of the model in all categories. CMC curve calculates the hit probability of Top-k, which comprehensively reflects the performance of the classifier. In our experiment, we show the scores of Rank-1, Rank-5 and Rank-10.


\begin{table*}[]
\caption{Experimental results of our method on RGBNT201 comparing with state-of-the-art methods in the case of one modality complete-missing.}
\begin{center}

\begin{tabular}{cccccccccl}
\toprule
\multicolumn{2}{c}{\multirow{2}{*}{\textbf{\textit{Methods}}}}              & \multicolumn{2}{c}{\textbf{No Missing}} & \multicolumn{2}{c}{\textbf{Missing NIR}} & \multicolumn{2}{c}{\textbf{Missing TIR}} & \multicolumn{2}{c}{\textbf{Missing NIR+TIR}} \\ 
\cmidrule(r){3-4} \cmidrule(r){5-6}  \cmidrule(r){7-8}  \cmidrule(r){9-10}
\multicolumn{2}{c}{}        & \textbf{mAP}         & \textbf{Rank-1}       & \textbf{mAP}            & \textbf{Rank-1}          & \textbf{mAP}            & \textbf{Rank-1}          & \textbf{mAP}              & \textbf{Rank-1}           \\ 
\cmidrule(r){1-2} \cmidrule(r){3-10} 
\multicolumn{1}{c}{\multirow{4}{*}{\textbf{\textit{Single-Modality}}}} & MLFN \cite{MLFN}  & 24.66          & 23.68          & 21.03          & 21.48           & 22.52          & 22.78           & 19.40            & 19.85            \\ 
\multicolumn{1}{c}{}                              & HACNN \cite{HACNN} & 19.34          & 14.71          & 16.78          & 13.50           & 13.44          & 11.23           & 12.34            & 10.43            \\ 
\multicolumn{1}{c}{}                              & OSNet \cite{OSNet}  & 22.12          & 22.85          & 17.90          & 18.44           & 20.43          & 21.03           & 16.76            & 17.23            \\ 
\multicolumn{1}{c}{}                              & CAL \cite{rao2021CAL}   & 25.63          & 26.30          & 23.05          & 23.58           & 21.35          & 22.40           & 20.52            & 21.63            \\ 
\cmidrule(r){1-2} \cmidrule(r){3-4} \cmidrule(r){5-6}  \cmidrule(r){7-8} \cmidrule(r){9-10} 
\multicolumn{1}{c}{\multirow{2}{*}{\textbf{\textit{Multi-Modality}}}}                   & PFNet \cite{Zheng2021Multi}  & 38.46          & 38.88          & 31.90          & 29.78           & 25.50          & 25.83           & 26.40            & 23.44            \\ 

                   & \textbf{DENet (Ours)}
&\textbf{ 42.41 }         & \textbf{42.23}          & \textbf{35.40}          & \textbf{36.82}           & \textbf{33.00}          & \textbf{35.40}           & \textbf{32.42}            & \textbf{29.20 }           \\ 
\bottomrule
\end{tabular}
\end{center}
\label{Tab:RGBNT201-diffMissing}
\end{table*}

\begin{table*}[]
\setlength{\abovecaptionskip}{0.2cm} 
\setlength{\belowcaptionskip}{-0.2cm}
\caption{Experimental results of our method on RGBNT201 are compared with state-of-the-art methods in the case of complete-missing of two modalities and no missing of modalities.}
\begin{center}

\begin{tabular}{cccccccccc}
\toprule
\multicolumn{2}{c}{\multirow{3}{*}{\textbf{\textit{Methods}}}}      & \multicolumn{4}{c}{\textbf{Missing NIR}}    & \multicolumn{4}{c}{\textbf{Missing TIR}}      \\ 
\cmidrule(r){3-6} \cmidrule(r){7-10}
\multicolumn{2}{c}{}          & \multicolumn{2}{c}{\textbf{RGB to TIR}}     & \multicolumn{2}{c}{\textbf{TIR to RGB}} & \multicolumn{2}{c}{\textbf{RGB to NIR}}     & \multicolumn{2}{c}{\textbf{NIR to RGB}} \\ 
\cmidrule(r){3-4}  \cmidrule(r){5-6}  \cmidrule(r){7-8}  \cmidrule(r){9-10}
\multicolumn{2}{c}{}                              & \textbf{mAP }  & \multicolumn{1}{c}{\textbf{Rank-1}} & \textbf{mAP}    & \textbf{Rank-1}    & \textbf{mAP}   & \multicolumn{1}{c}{\textbf{Rank-1}} & \textbf{mAP}      & \textbf{Rank-1}     \\ 
\cmidrule(r){1-2} \cmidrule(r){3-4}  \cmidrule(r){5-6}  \cmidrule(r){7-8}  \cmidrule(r){9-10}
\multicolumn{1}{c}{\multirow{3}{*}{\textbf{\textit{Cross-Modality}}}} & HC loss \cite{HCloss} & 16.74 & \multicolumn{1}{c}{14.25}  & 16.53          & 19.32          & 20.54 & \multicolumn{1}{c}{22.19}  & 21.80         & 22.93          \\ 
\multicolumn{1}{c}{}                             & DDAG \cite{Ye2020DDAG}    & 18.09 & \multicolumn{1}{c}{14.79}  & 20.01          & 18.05          & 21.39 & \multicolumn{1}{c}{20.37}  & 24.76         & 21.66          \\  
\multicolumn{1}{c}{}                             & MPANet \cite{Wu2021MPANet}  & 19.00 & \multicolumn{1}{c}{20.82}  & 20.28          & 23.14          & 20.04 & \multicolumn{1}{c}{30.96}  & 26.03         & 26.54          \\ 
\cmidrule(r){1-2} \cmidrule(r){3-4}  \cmidrule(r){5-6}  \cmidrule(r){7-8}  \cmidrule(r){9-10}
\multicolumn{1}{c}{\textbf{\textit{Ours}}}                   & \textbf{DENet}
& \multicolumn{4}{c}{mAP: \textbf{35.4} \quad \quad Rank-1: \textbf{36.82}}                            & \multicolumn{4}{c}{mAP: \textbf{33.00} \quad \quad Rank-1: \textbf{35.40}}                          \\ 
\bottomrule
\end{tabular}
\end{center}
\label{Tab:RGBNT201-missingOne}
\end{table*}

\begin{table*}[]
\setlength{\abovecaptionskip}{0.2cm} 
\setlength{\belowcaptionskip}{-0.2cm}
\caption{Comparison results between our method DENet and Baseline model when using data of different modal combinations on RGBNT100 dataset.}
\begin{center}
\setlength{\tabcolsep}{2mm}{
\begin{tabular}{ccccccccccccc}
\toprule
\multirow{2}{*}{\textbf{\textit{Methods}}}   & \multicolumn{3}{c}{\textbf{No Missing}} & \multicolumn{3}{c}{\textbf{Missing NIR}} & \multicolumn{3}{c}{\textbf{Missing TIR}} & \multicolumn{3}{c}{\textbf{Missing NIR+TIR}} \\ 
\cmidrule(r){2-4} \cmidrule(r){5-7} \cmidrule(r){8-10} \cmidrule(r){11-13}
& \textbf{mAP}       & \textbf{Rank-1}    & \textbf{Rank-5}     & \textbf{mAP }       & \textbf{Rank-1}   & \textbf{Rank-5}       &\textbf{ mAP}         & \textbf{Rank-1}  & \textbf{Rank-5}    & \textbf{mAP }     & \textbf{Rank-1} & \textbf{Rank-5}\\ 
\cmidrule(r){1-1} \cmidrule(r){2-4} \cmidrule(r){5-7} \cmidrule(r){8-10} \cmidrule(r){11-13}
Baseline       & 62.1       & 87.1   & 89.0    & 58.9        & 85.5   & 87.6     & 53.8      & 79.2   & 82.0    & 46.5      & 70.0   &74.2    \\ 
\textbf{DENet}          & \textbf{68.1}       &\textbf{89.2}  &\textbf{91.1}      
& \textbf{62.0}       & \textbf{85.5}     & \textbf{88.1} 
&\textbf{56.0}      & \textbf{80.9}      & \textbf{84.5}
& \textbf{50.1}      & \textbf{74.2}       & \textbf{78.0} \\ 
\bottomrule
\end{tabular}}
\end{center}
\label{Tab:RGBNT100-1}
\end{table*}


\subsection{Evaluation on RGBNT201 Dataset}
In order to verify the effectiveness of DENet in processing partial-missing multi-modality data, we implement several groups of comparative experiments on RGBNT201 dataset according to the increasing number of missing modalities.

\noindent
{\bf No modality missing.}
To verify the generality of the proposed method for multi-modality person Re-ID, we compare our method with the state-of-the-art single and multi modality methods in Table~\ref{Tab:RGBNT201-diffMissing}. 
Firstly, we compare our DENet with four single modality methods, including MLFN \cite{MLFN}, HACNN \cite{HACNN}, OSNet \cite{OSNet}, and CAL \cite{rao2021CAL}. 
For the single modality method, we expand the network into three branches, and concatenate the obtained features as the representation for Re-ID task. Without missing, we put the obtained features into the enhancement module to have a complete directed enhance. 
Secondly, we compare with multi-modality Re-ID methods PFNet \cite{Zheng2021Multi} in complete setting. 
Clearly, our DENet significantly beats both the existing multi-modality Re-ID methods PFNet and these single modality methods, which demonstrates the effectiveness of the proposed method in complete multi-modality Re-ID.

\noindent
{\bf One modality complete-missing.}
In the case with one modality complete-missing, which means one fixed modality is missing for all the samples (taking TIR or NIR as example), we carry out a series of comparative experiments on RGB-NIR and RGB-TIR settings respectively.
In this case, we also compare our method with four single modality methods mentioned above. We extend the single modality network into two branches and concatenate the two features of existing modalities as the representation. 
Additionally, we evaluate the multi-modality Re-ID method PFNet \cite{Zheng2021Multi} in the fashion of missing modality. The result are shown in Table~\ref{Tab:RGBNT201-diffMissing}.
Furthermore, we construct the RGBNT201 dataset into cross-modality setting by query one modality from the other modality gallery.  
Thus, we evaluate three cross-modality Re-ID methods HC loss \cite{HCloss}, DDAG \cite{Ye2020DDAG} and MPANet \cite{Wu2021MPANet} for comparison. The results are shown in Table~\ref{Tab:RGBNT201-missingOne}.
Cross-modality methods mainly focus on reducing the modality heterogeneity while ignoring the complementarity in different modalities, thus present generally poor accuracy in both mAP and Rank-1.
Single modality methods integrate the multi-modality information between the two existing modalities by simple concatenation, thus still achieves relatively poor performance.
By considering the heterogeneity of multi-modality data and recovering the modality missing by cross-modality transfer, PFNet achieves large improvement comparing to the cross-modality and single modality methods.
However, due to the low constraints of the recovery process and insufficient attention to modality-specific features, which limits its performance.
Benefit from the necessary restoration of missing modalities by the feature transformation module, as well as the useful information mining and interaction of dynamic feature enhancement, our DENet significantly boosts the performance, which evidences the effectiveness of our multi-modality while handling the partial multi-modality person re-identification.

\noindent
{\bf Two modality complete-missing.}
In the case of missing two modalities (taking TIR and NIR as example), we take RGB image as input and compare our method with the above three single-modality Re-ID method and the multi-modality Re-ID method PFNet with missing modalities. 
Consistently shown in Table~\ref{Tab:RGBNT201-diffMissing}, our method significantly beats both single and multi-modality methods, which demonstrates the effectiveness of the proposed method in partial multi-modality Re-ID.


\begin{table}[]
\setlength{\abovecaptionskip}{0.2cm} 
\setlength{\belowcaptionskip}{-0.2cm}
\caption{Comparison with state-of-the-art single-modality and multi-modality Re-ID methods on RGBNT100.}
\begin{center}
\setlength{\tabcolsep}{1mm}{
\begin{tabular}{cccccc}
\toprule
\multicolumn{2}{c}{\textbf{\textit{Methods}}}                                      & \textbf{mAP}  & \textbf{Rank-1} & \textbf{Rank-5} & \textbf{Rank-10} \\ 
\cmidrule(r){1-2}\cmidrule(r){3-6}
\multicolumn{1}{c}{\multirow{2}{*}{\textbf{\textit{Single-Modality}}}} & DMML~\cite{DMML}        & 58.5 & 82.0   & 85.1   & 86.2    \\ 

\multicolumn{1}{c}{}                              & Circle Loss~\cite{Circle-Loss} & 59.4 & 81.7   & 83.7   & 85.2    \\ 
\cmidrule(r){1-2}\cmidrule(r){3-6}
\multicolumn{1}{c}{\multirow{2}{*}{\textbf{\textit{Multi-Modality}}}}                   & HAMNet~\cite{Li2020HAMNet}      & 64.1 & 84.7   & 88.0   & 89.4    \\
\multicolumn{1}{c}{} &{\textbf{DENet (Ours)}}           & \textbf{68.1} & \textbf{89.2}   & \textbf{91.1}   & \textbf{92.0}    \\ 
\bottomrule
\end{tabular}}
\end{center}
\label{Tab:RGBNT100-2}
\end{table}


\subsection{Evaluation on RGBNT100 Dataset}
To verify the generality of our method, we further validate our DENet on the multi-spectral vehicle Re-ID dataset RGBNT100 \cite{Li2020HAMNet}.  

Firstly, we evaluate our DENet on the RGBNT100 dataset in the case of diverse modality missing comparing with the baseline, as shown in Table~\ref{Tab:RGBNT100-1}.
It can be observed that DENet achieves the best performance when there is no modality missing. 
The performance degrades when one modality (NIR or TIR) missing, and even worse when both NIR and TIR are missing.
This evidence that we can use the complementary information provided by multi-modality images to force the network to learn more discriminant feature representations, so as to obtain more robust performance.
Moreover, by introducing Cross-Modality Feature Transformation (CMFT) and Dynamic Enhancement Module (DEM), our DENet significantly improves the baseline in all the cases. This verifies the effectiveness of our method by applying our model to the multi-modality vehicle dataset RGBNT100.
%


In addition, we compare our method DENet with the multi-modality vehicle Re-ID method HAMNet and the advanced single-modal Re-ID method DMML\cite{DMML} and Circle Loss\cite{Circle-Loss}. Specifically, we extend the single-modal models to three branches, followed by the feature concatenation to obtain the final vehicle feature representation, then to facilitate the multi-modality Re-ID task.
As shown in Table~\ref{Tab:RGBNT100-2}, DENet sigfinicantly beats these comparative methods, which evidences the effectiveness of our method while handling multi-modality information.


\begin{table}[]
\setlength{\abovecaptionskip}{0.2cm} 
\setlength{\belowcaptionskip}{-0.2cm}
\caption{Ablation study of partial multi-modality Re-ID on RGBNT201 with missing rate $\eta = 0.25$. "\checkmark" indicates that the component is included.}
\begin{center}
\setlength{\tabcolsep}{2mm}{
\begin{tabular}{ccccccc}
\toprule
\textbf{$L_{rec}$} & \textbf{$L_{sim}$} & $DEM$ & \textbf{mAP}  & \textbf{Rank-1} & \textbf{Rank-5} & \textbf{Rank-10} \\ 
\cmidrule(r){1-3}\cmidrule(r){4-7}
   -    &   -   &  -    & 25.95    & 23.12      & 36.13      & 45.20    \\
\cmidrule(r){1-3}\cmidrule(r){4-7}
\checkmark    &  -    &  -    & 30.52    & 27.91      & 41.40      & 51.73    \\
   -  & \checkmark    &   -   & 31.05 & 28.60   & 45.94   & 56.85    \\
   -  &   -   & \checkmark    & 31.42 & 29.10   & 44.98   & 53.82    \\
  \cmidrule(r){1-3}\cmidrule(r){4-7}
\checkmark    &   -   & \checkmark    & 32.93 & 31.64   & 47.67   & 59.90    \\
  -   & \checkmark    & \checkmark    & 31.20 & 32.58   & 52.25   & 61.17    \\
\checkmark    & \checkmark    &  -    & 33.40 & 35.52   & 52.40   & 62.83    \\
\cmidrule(r){1-3}\cmidrule(r){4-7}
\checkmark    & \checkmark    & \checkmark    & \textbf{34.30} & \textbf{40.18}  & \textbf{57.42}   & \textbf{65.90}    \\ 
\bottomrule
\end{tabular}}
\end{center}
\label{Tab:ablation_study}
\end{table}

\begin{figure*}[t]
\setlength{\abovecaptionskip}{2pt} 
\setlength{\belowcaptionskip}{-2pt}
\begin{center}
\includegraphics[width=1\linewidth]{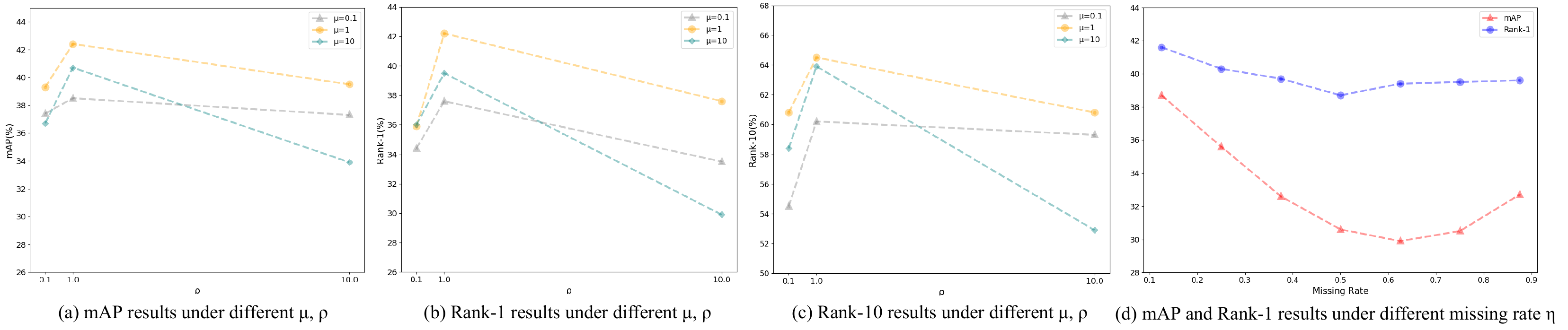}
\end{center}
\caption{
The parameter analysis on RGBNT201 dataset shows the influence of different coefficient settings on the accuracy of the learned model.
}
\label{fig:Parameter_Analysis}
\end{figure*}


\begin{table}[]
\setlength{\abovecaptionskip}{0.2cm} 
\setlength{\belowcaptionskip}{-0.2cm}
\caption{Comparative experiment between CMFT module and GAN-based recovery approaches on RGBNT201 with missing rate $\eta = 0.25$.}
\begin{center}
\setlength{\tabcolsep}{3mm}{
\begin{tabular}{ccccc}
\toprule
\textbf{\textit{Methods}}             & \textbf{mAP}   & \textbf{Rank-1} & \textbf{Rank-5} & \textbf{Rank-10} \\ 
\cmidrule(r){1-1} \cmidrule(r){2-5}
Image Generation    & 30.70 & 32.61  & 41.45  & 50.74   \\
Feature Generation & 31.62 & 35.30  & 47.25  & 57.73   \\
\textbf{DENet (Ours)}              & \textbf{34.30} & \textbf{40.18}  & \textbf{57.42}  & \textbf{65.90}   \\ 
\bottomrule
\end{tabular}}
\end{center}
\label{Tab:Compare_CMFT_1}
\end{table}


\begin{table}[h]
\setlength{\abovecaptionskip}{0.2cm} 
\setlength{\belowcaptionskip}{-0.2cm}
\caption{Comparative experiment of dynamic enhancement and other enhancement strategies on RGBNT201 datatset.}
\begin{center}
\setlength{\tabcolsep}{1.5mm}{
\begin{tabular}{cccccc}
\toprule
 \textbf{\textit{Missing Modality}}  & \textbf{\textit{Methods}}  & \textbf{mAP}  & \textbf{Rank-1} & \textbf{Rank-5} & \textbf{Rank-10} \\ 
\cmidrule(r){1-2} \cmidrule(r){3-6}
\multicolumn{1}{c}{\multirow{2}{*}{TIR}}    &  Fixed & 30.00 & 32.51  & 53.62  & 62.10   \\
\multicolumn{1}{c}{}                                &  \textbf{Ours} & \textbf{33.00} & \textbf{35.40} & \textbf{55.02}  & \textbf{62.90}  \\  
\cmidrule(r){1-2} \cmidrule(r){3-6}
\multicolumn{1}{c}{\multirow{2}{*}{NIR}}    &Fixed \quad & 34.85 & 34.60  & 48.82  & 57.81   \\
\multicolumn{1}{c}{}                                &\textbf{Ours}  & \textbf{35.40} & \textbf{36.82}  & \textbf{53.60}  & \textbf{64.10}   \\ 
\cmidrule(r){1-2} \cmidrule(r){3-6}
\multicolumn{1}{c}{\multirow{2}{*}{NIR+TIR}} &  Fixed  & 31.10 & 26.91  & 42.00  & 52.26   \\
\multicolumn{1}{c}{}                                & \textbf{Ours}  & \textbf{32.42} & \textbf{29.20}  & \textbf{45.26}  & \textbf{53.38}   \\ 
\cmidrule(r){1-2} \cmidrule(r){3-6}
\multicolumn{1}{c}{\multirow{3}{*}{None}}       & Fixed            &  39.25     & 36.64       &  52.37   &  62.80       \\
\multicolumn{1}{c}{}                                & single-direction & 35.02     & 33.14      &  47.10      &  57.95      \\
\multicolumn{1}{c}{}                                & \textbf{Ours}             & \textbf{42.41} &\textbf{42.23}  & \textbf{55.30}  & \textbf{64.52}   \\ 
\bottomrule
\end{tabular}}
\end{center}
\label{Tab:Compare_DEM}
\end{table}

\subsection{Ablation Study} 
To verify the contribution of each component, reconstruction loss $L_{rec}$, similarity loss $L_{sim}$ and the Dynamic Enhancement Module (DEM) in our method, we conduct the following ablation study on RGBNT201 with missing rate $\eta = 0.25$. 
As shown in Table~\ref{Tab:ablation_study}, each component plays indispensable role in our method. 
In the case of partial multi-modality data, compared with no conversion constraints, our dynamic enhancement module can maintain the authenticity of the restored features as much as possible, thus can effectively deal with the missing data.
The performance can be further improved by using the dynamic feature enhancement module for modal interaction under different missing state.


\noindent
\textbf{Evaluation on CMFT}.
The proposed cross-modality feature transformation (CMFT) aims to recover the features of missing modalities. To further verify the effectiveness of this module, we first compare it with other conversion approaches including GAN-based image generation and feature generation methods on RGBNT201 dataset.
These two comparison methods use the training strategy of Generative Adversarial Networks (GAN) to generate false missing images and features respectively, so as to replace our CMFT module.
As can be seen from Table~\ref{Tab:Compare_CMFT_1}, the CMFT module can achieve superior performance than the other two compared ways. One of the main reasons is that the training of GAN needs a large amount of training data and requires better synchronization between the generator and the discriminator. The training process is prone to non convergence, resulting in unstable results, which limits the improvement on multi-modality person Re-ID.


\noindent
\textbf{Evaluation on DEM}.
In order to further verify the effectiveness of the proposed dynamic enhancement module (DEM), we design the comparative experiments of fixed enhancement in diverse modality missing cases. Fixed enhancement strategy means that only the RGB is used to enhance NIR and TIR respectively, while the TIR and NIR will not enhance other modalities, given that they may be missing. 
As shown in Table~\ref{Tab:Compare_DEM}, the flexible use of dynamic cutting strategy for dynamic enhancement can obtain better performance than fixed enhancement. The main reason is that our strategy considers the availability of existing features and restored features reasonably, rather than treating them equally. 
Furthermore, in the complete setting with no modality missing, we compare our method with the single-direction cyclic enhancement strategy, which takes RGB, NIR and TIR as nodes and R2N, N2T and T2R as edges. It can be seen form Table~\ref{Tab:Compare_DEM}, either single-direction cyclic enhancement or fixed enhancement works overshadowed than the dual-direction full enhancement in the DEM in the case of data integrity.


\begin{figure}[t]
\setlength{\abovecaptionskip}{2pt} 
\setlength{\belowcaptionskip}{-2pt}
\begin{center}
\includegraphics[width=1\linewidth]{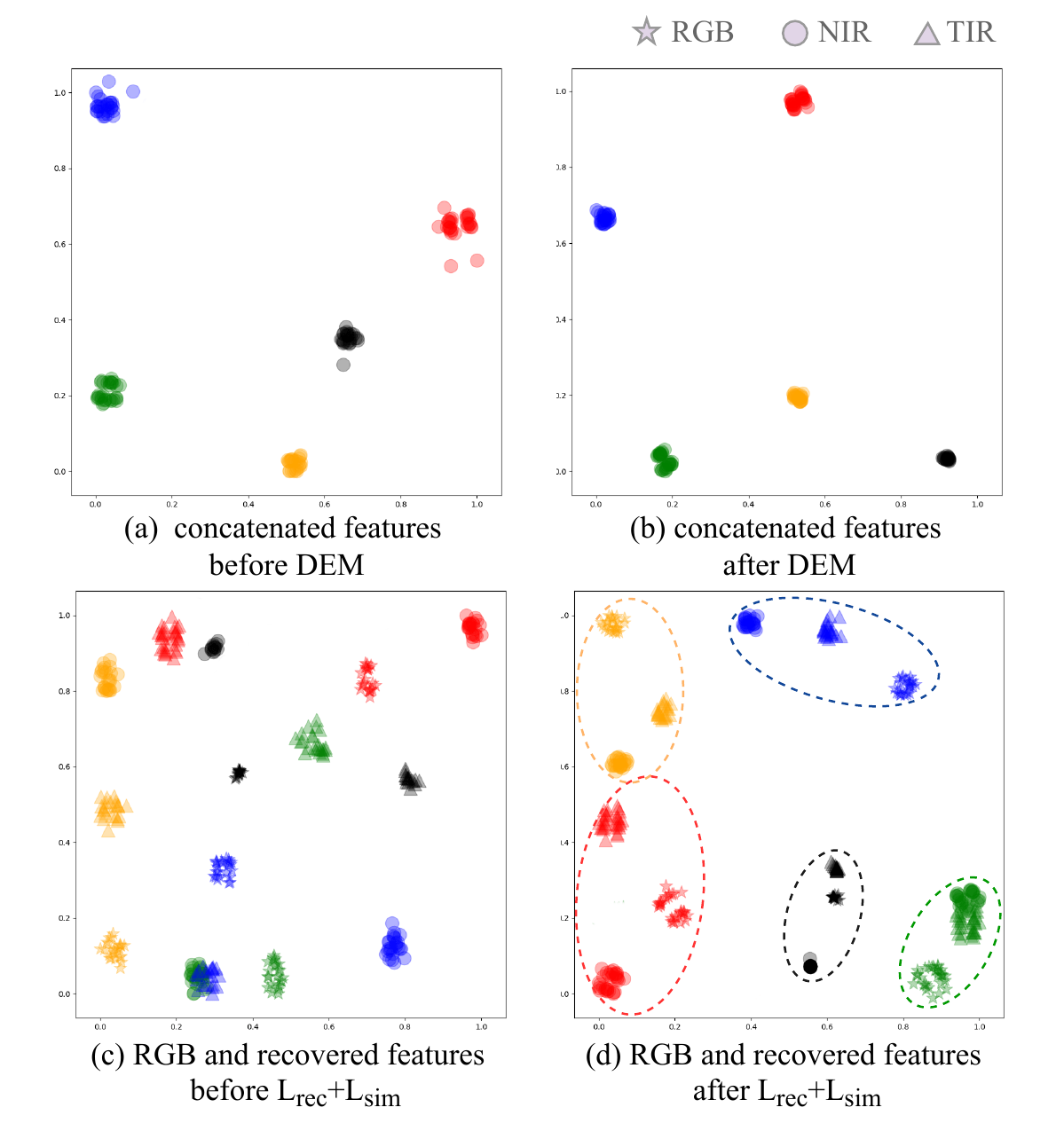}
\end{center}
\caption{
t-SNE visualization \cite{2008Visualizing} of the final features and three modality features. Different colors and shapes represent different IDs and modalities.
}
\label{fig:visual}
\end{figure}


\begin{figure}[t]
\setlength{\abovecaptionskip}{2pt} 
\setlength{\belowcaptionskip}{-2pt}
\begin{center}
\includegraphics[width=1\linewidth]{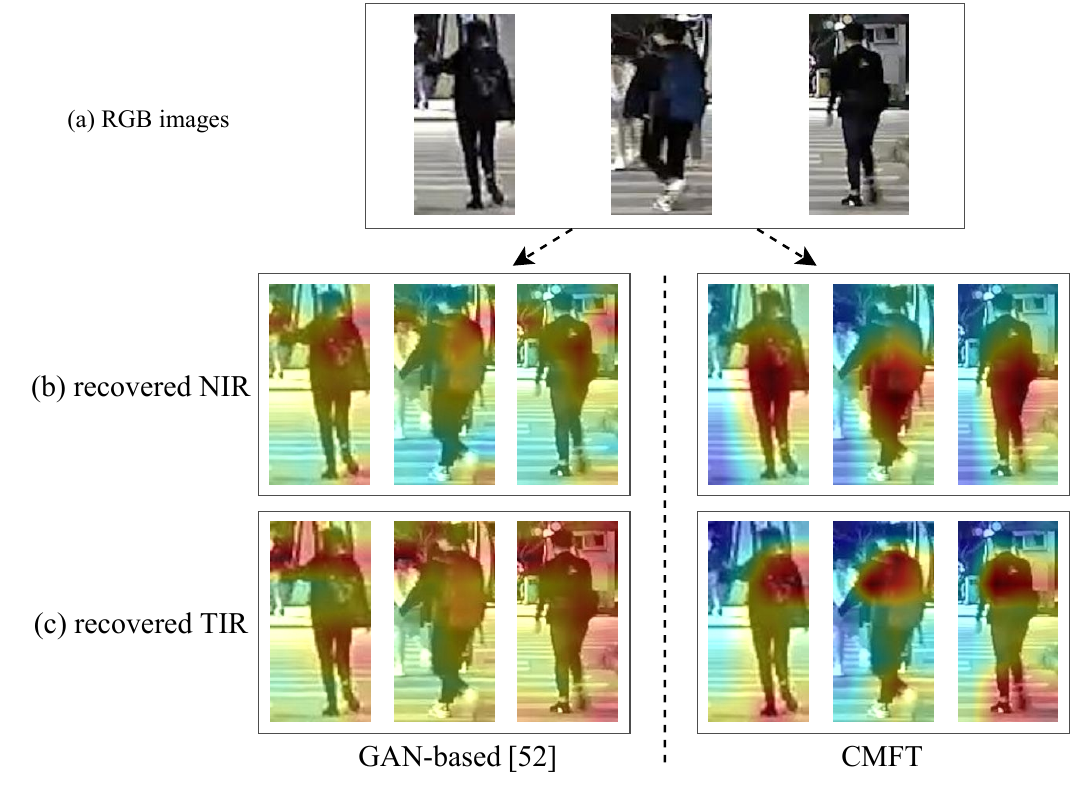}
\end{center}
\caption{
Comparison of the Class Activation Map (CAM) of recovered (b) NIR features and (c) TIR features overlaid on the existing (a) RGB images.
}
\label{fig:featuremap}
\end{figure}


\begin{figure*}[t]
\setlength{\abovecaptionskip}{2pt} 
\setlength{\belowcaptionskip}{-2pt}
\begin{center}
\includegraphics[width=1\linewidth]{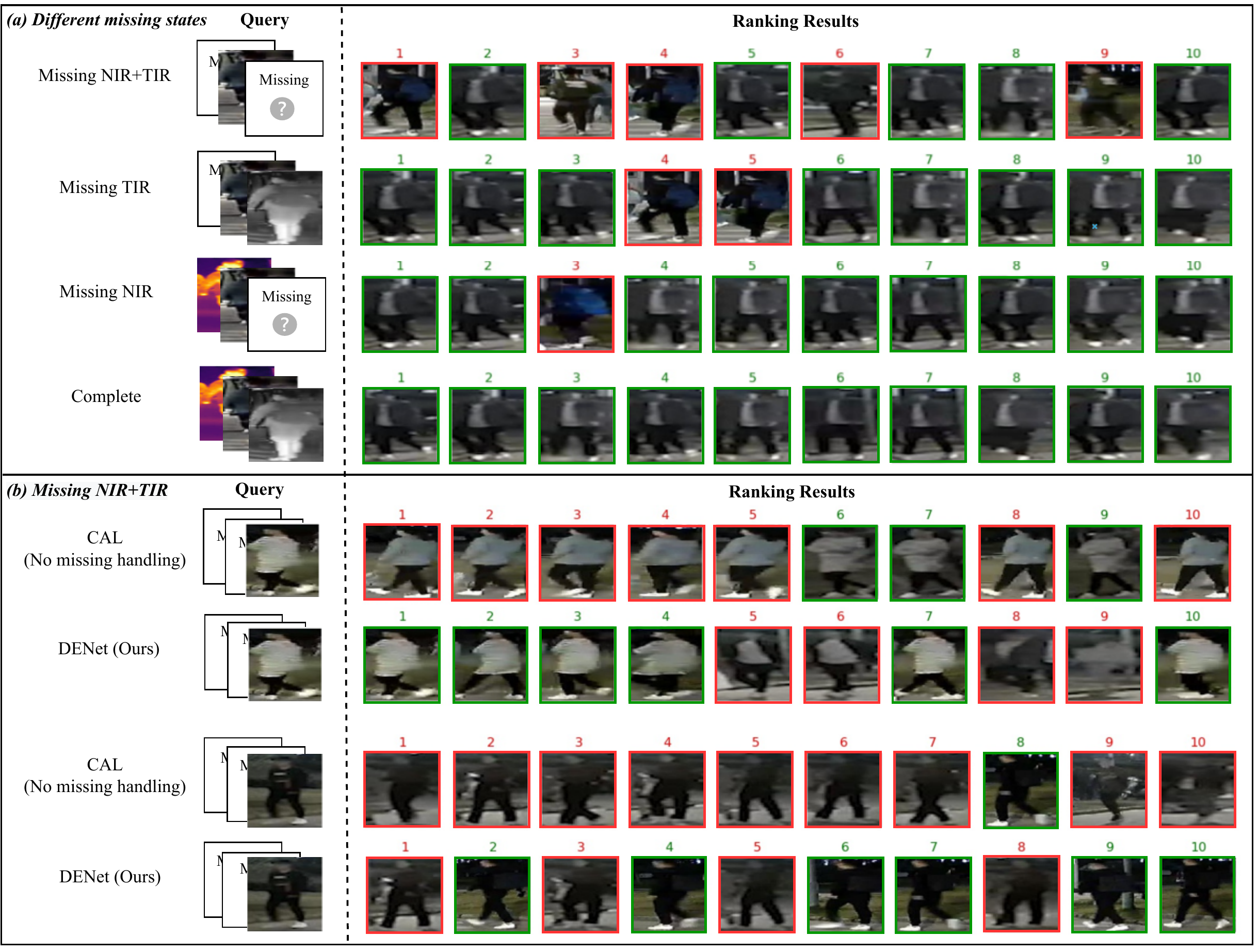}
\end{center}
\caption{
(a) The ranking results of an example query in different missing scenarios. (b) The ranking results of our DENet comparing with the state-of-the-art method when missing NIR and TIR modalities. The green and red boxes indicate the correct and false matching respectively.
}
\label{fig:ranking}
\end{figure*}


\subsection{Parameter Analysis}

There are three important parameters in our method, the influence of the values of parameters $\rho$ and $\mu$, and the missing rate $\eta$.
First, we evaluate the influence of $\rho$ and $\mu$ on training accuracy, as shown in Fig.~\ref{fig:Parameter_Analysis} (a)-(c).
Our method is relatively robust under the setting of $\mu$ = 1, and achieves the best performance when $\rho$ = 1 and $\mu$ = 1. Therefore, in other experiments, we set $\rho$ = 1 and $\mu$ = 1 for comparison.
Second, to explore the capability of the proposed method while handling performance with the more realistic partial multi-modality Re-ID, we conduct experiments with different missing rates on RGBNT201 dataset. 
Specifically, we randomly remove $M$ images (could be any modality for each multi-modality/triplet sample) from $N$ quegallery images in the test stage. Fig.~\ref{fig:Parameter_Analysis} (d) demonstrates the performance against the missing rate $\eta = M/N$. 
It can be seen that the two evaluation indicators first decreased with the increase of the missing rate, and turned to an upward trend near the missing rate of 0.6 and 0.5 respectively. The main reason may be that the image quality of some NIR and TIR is not clear, and the features extracted from the original real images are not recognized as the features converted from RGB, which also reflects the effectiveness of the feature transformation module.


\subsection{Visualization}
To illustrate the effectiveness of the proposed dynamic enhancement module (DEM) on RGBNT201 dateset, we randomly visualize the feature distribution of five identities from the test set by t-SNE \cite{2008Visualizing}, where different colors and shapes represent different identities and modalities respectively.
From Fig.~\ref{fig:visual} (a)-(b), we can see that our dynamic enhancement module can explore the supplementary information of different modalities and maintain effective correlation, making the feature distribution in Fig.~\ref{fig:visual} (b) more compact. 
In addition, benefit from the two loss functions $L_{rec}$ and $L_{sim}$ in the CMFT module, we can obtain more realistic recovered feature representation as shown in Fig.~\ref{fig:visual} (c)-(d). 
The transformation significantly alleviate the modality differences, thus the features with same identity but different modalities cluster better than the original.

Secondly, since one of the key points of multi-modality person Re-ID is to improve the discriminability of the features, we further explicitly show the effectiveness of the cross-modality feature transformation compared to the GAN-based generation method \cite{Cross-Modality-GAN} in terms of missing modality recovery. 
We apply Grad-Cam \cite{2017Grad-Cam} on the generated/recovered feature maps to visualize the class activation maps (CAMs) and overlay them on the original RGB images as shown in Fig.~\ref{fig:featuremap}.
We observe that the GAN-based feature generation method is more heavily disturbed by the background and does not pay enough attention to the target. 
In contrast, the features recovered by our method better emphasize the most discriminative regions required for classification. The stronger the feature discriminability, the better the network performance.


Finally, we visualize the several ranking results of our model retrieved on the RGBNT201 dataset, as shown in Fig.~\ref{fig:ranking}. For clarity, we only show the RGB images in the ranking results. 
Fig.~\ref{fig:ranking} (a) demonstrates the ranking results of an example query in different missing scenarios. We can see that the best retrieval results achieve when the multi-modality data is complete, and the performance is relatively the worst when both NIR and TIR are missing. This reflects that the complementarity of multi-modality data plays an effective role in improving network performance.
Fig.~\ref{fig:ranking} (b) further demonstrates the results of our DENet comparing with the state-of-the-art CAL~\cite{rao2021CAL} when missing NIR and TIR modalities. It can be seen that comparing with the method without special handling of modality missing, our DENet complements the missing information through the CMFT module and then selectively interacts through the DEM, which can achieve much more robust Re-ID performance.


\section{Conclusion}

In this paper, we propose a novel partial multi-modality Re-ID method DENet to cope with the missing problem as well as improve the multi-modality representation. First, we introduce the cross-modality feature transformation module to recover the representation of missing data. In addition, we design a dynamic enhancement module with a complete directed graph, and then cut relative edges according to the missing state dynamically. It improves the representation ability of multiple modalities under changeable missing scenarios. Comprehensive experiment results on the RGBNT201 and RGBNT100 datasets demonstrate the performance of our method and validate that our DENet is better compatible with challenging modality-missing environments. In the future, we will focus on more complex situations to achieve more robust partial multi-modality Re-ID performance.



\bibliographystyle{IEEEtran}
\bibliography{trans}


\end{document}